\documentclass[sn-mathphys]{sn-jnl}

\jyear{2022}%

\usepackage{array,epsfig,rotating}
\usepackage{graphicx}
\usepackage{subfigure}
\usepackage{booktabs} 
\usepackage{amsmath}
\usepackage{amssymb}
\usepackage{mathtools}
\usepackage{amsthm}
\usepackage{bm}
\usepackage{mathrsfs}
\usepackage{natbib}
\usepackage{hyperref}

\catcode`\|=12

\theoremstyle{thmstylethree}%
\newtheorem{theorem}{Theorem}
\newtheorem{lemma}{Lemma}

\newtheorem{assumption}{Assumption}
\theoremstyle{remark}
\newtheorem{remark}{Remark}
\begin{document}

\title[Estimating Mixed-Memberships by Mixed-SLIM]{Estimating Mixed-Memberships Using the Symmetric Laplacian Inverse Matrix}


\author[1]{\fnm{Huan} \sur{Qing}}\email{qinghuan@cumt.edu.cn}

\author*[2]{\fnm{Jingli} \sur{Wang}}\email{jlwang@nankai.edu.cn}

\affil[1]{\orgdiv{School of Mathematics}, \orgname{China University of Mining and Technology},  \city{Xuzhou}, \postcode{221116}, \state{Jiangsu, \country{China}}}

\affil[2]{\orgdiv{School of Statistics and Data Science, KLMDASR, LEBPS, and LPMC}, \orgname{Nankai University},  \city{Tianjin}, \postcode{30071}, \state{Tianjin}, \country{China}}


\abstract{ Mixed membership community detection is a challenging problem. In this paper, to detect mixed memberships, we propose a new method Mixed-SLIM which is a spectral clustering method on the symmetrized Laplacian inverse matrix under the degree-corrected mixed membership model. We provide theoretical bounds for the estimation error on the proposed algorithm and its regularized version under mild conditions. Meanwhile, we provide some extensions of the proposed method to deal with large networks in practice. These Mixed-SLIM methods outperform state-of-art methods in simulations and substantial empirical datasets for both community detection and mixed membership community detection problems.}




\maketitle

\section{Introduction}

The development of the Internet not only changes people's lifestyles but also produces and records a large number of network structure data. Therefore, networks are often associated with our life, such as friendship networks and social networks, and they are also essential in science, such as biological networks \citep{2002Food}, information networks \citep{Newman2004} and social networks \cite{pizzuti2008ga, Scoot2014}. To analyze networks, many researchers present them in a form of a graph in which subjects/individuals are presented by nodes, and the relationships are measured by the edges, directions of edges, and weights \cite{fortunato2010community, fortunato2016community}. Some authors consider `pure' networks in which each node at most belongs to one community/cluster, and in each community, the nodes which have similar proprieties or functions are more likely to be linked with each other than random pairs of nodes \citep{RSC, SCORE}. While few networks can be deemed as `pure' in our real life. In a network, if some nodes are potentially belonging to two or more communities at a time, the network is known as `mixed membership' \citep{mixedSCORE, OCCAM, SPACL}. Compared with pure networks, mixed membership networks are more realistic. In this paper, we focus on the problem of community detection for mixed membership networks.

The stochastic blockmodel (SBM) \citep{SBM} is one of the most used models for community detection in which all nodes in the same community are assumed to have equal expected degrees. Some recent developments of SBM can be found in \citep{abbe2017community} and references therein. Since in empirical network data sets, the degree distributions are often highly inhomogeneous across nodes, a natural extension of SBM is proposed: the degree-corrected stochastic block model (DCSBM) \citep{DCSBM} which allows the existence of degree heterogeneity within communities. DCSBM is widely used for community detection for non-mixed membership networks \citep{zhao2012consistency, SCORE,cai2015robust, chen2018convexified, chen2018network, ma2021determining}. \cite{MMSB} constructed a mixed membership stochastic blockmodel (MMSB) which is an extension of SBM by letting each node have different weights of membership in all communities. However, in MMSB, nodes in the same communities still share the same degrees. To overcome this shortcoming, \cite{mixedSCORE} proposed a degree-corrected mixed membership (DCMM) model. DCMM model allows that nodes for the same communities have different degrees and some nodes could belong to two or more communities, thus it is more realistic and flexible. In this paper, we design community detection algorithms based on the DCMM model.

In this paper, we extend the symmetric Laplacian inverse matrix (SLIM)  method \citep{SLIM} to mixed membership networks and call this proposed method as mixed-SLIM. As mentioned in \cite{SLIM}, the idea of using the symmetric Laplacian inverse matrix to measure the closeness of nodes comes from the first hitting time in a random walk.  \cite{SLIM} combined the SLIM with the spectral method based on DCSBM for community detection.  And the SLIM method outperforms state-of-art methods in many real and simulated datasets. Therefore, it is worth modifying this method to mixed membership networks. Numerical results of simulations and substantial empirical datasets in Section \ref{sec5} show that our proposed Mixed-SLIM indeed enjoys satisfactory performances when compared to the benchmark methods for both community detection problem and mixed membership community detection problem.
\section{Degree-corrected mixed membership model}\label{sec2}
First, we introduce some notations. $\|\cdot\|_{F}$ for a matrix denotes the Frobenius norm,  $\|\cdot\|$ for a matrix denotes the spectral norm,  and $\|\cdot\|$ for a vector denotes the $l_{2}$-norm. For convenience, when we say ``leading eigenvalues'' or ``leading eigenvectors'', we are comparing the \emph{magnitudes} of eigenvalues and their respective eigenvectors with unit-norm. For any matrix or vector $x$, $x'$ denotes the transpose of $x$. For any matrix $X$, we simply use $Y=\mathrm{max}(0, X)$ to represent $Y_{ij}=\mathrm{max}(0, X_{ij})$ for any $i,j$.

Assume we have an undirected, un-weighted, and no-self-loops network $\mathcal{N}$. Let $A$ be its adjacency matrix such that $A_{ij}=1$ if there is an edge between node $i$ and $j$, $A_{ij}=0$ otherwise, for $i,j=1,...,n$. We also assume that there are $K$ disjoint blocks $V^{(1)}, V^{(2)}, \ldots, V^{(K)}$ where $K$ is the number of clusters/communities which is assumed to be known in this paper.

In this paper, we consider the degree-corrected mixed membership (DCMM) model \citep{mixedSCORE}. For a mixed membership network, nodes could belong to multiple clusters. To measure how likely each node belongs to a certain community, DCMM assumes that node $i$ belongs to cluster $V^{(k)}$ with probability $\pi_{i}(k)$, i.e.,
\begin{align*}
	\mathrm{Pr}(i \in V^{(k)})=\pi_{i}(k), \qquad 1\leq k\leq K, 1\leq i\leq n,
\end{align*}
and $\sum_{k=1}^{K}\pi_{i}(k)=1$, where $\pi_{i}=(\pi_{i}(1), \pi_{i}(2), \ldots, \pi_{i}(K))$ which is known as Probability Mass Function (PMF). We call node $i$ `pure' if there is only one element of $\pi_i$ is 1, and all other $K-1$ entries are 0; and call node $i$ `mixed' otherwise. We call $\underset{1\leq k\leq K}{\mathrm{max}}\pi_{i}(k)$ as the \textit{purity} of node $i$.

Then the adjacency matrix can be modeled by the DCMM model. First model the degree heterogeneity by a positive vector $\theta=(\theta(1), \ldots, \theta(n))'$. Then if we know that node $i\in V^{(k)}$ and node $ j\in V^{(l)}$, for any $i,j = 1,...n,$ and $k,l = 1, ..., K$, we have the following conditional probability
\begin{equation*}
	Pr(A(i,j)=1 | i \in V^{(k)}, j\in V^{(l)})=\theta(i)\theta(j)P(k,l),
\end{equation*}
 where $P$ is  a $K\times K$ symmetric non-negative matrix  (called mixing matrix in this paper) such that $P$ is non-singular, irreducible, and $P(i,j)\in [0,1] \mathrm{~for~}1\leq i,j\leq K$.
For $1\leq i<j\leq n$, $A(i,j)$ are Bernoulli random variables that are independent of each other, satisfying
\begin{align*}
\mathrm{Pr}(A(i,j)=1)=\theta(i)\theta(j)\sum_{k=1}^{K}\sum_{l=1}^{K}\pi_{i}(k)\pi_{j}(l)P(k,l).
\end{align*}
Introduce the $n\times K$ membership matrix $\Pi$ such that the $i$-th row of $\Pi$ (denoted as $\Pi_{i}$) is $\pi_{i}$ for all $i\in \{1,2,\ldots, n\}$. Let $E[A]=\Omega$ such that $\Omega(i,j)=\mathrm{Pr}(A(i,j)=1), 1\leq i<j\leq n$. Then we have
\begin{align*}
	\Omega=\Theta \Pi P \Pi' \Theta,
\end{align*}
where $\Theta$ is an $n\times n$ diagonal matrix whose $i$-th diagonal entry is $\theta_{i}$ for $1\leq i\leq n$.

Given $(n, P, \Theta, \Pi)$, we can generate a random adjacency matrix $A$ under DCMM. For convenience, we denote the DCMM model as $DCMM(n, P, \Theta, \Pi)$  in this paper. For mixed membership community detection, the chief aim is to estimate $\Pi$ with given $A$ and known $K$.
\section{Methodology}\label{sec3}
In this section, we first introduce the main algorithm mixed-SLIM which can be taken as a natural extension of the SLIM \citep{SLIM} to the mixed membership community detection problem. Then we discuss the choice of some tuning parameters in the proposed algorithm.
\subsection{Algorithm: mixed-SLIM}

Roughly, the main idea is that the estimation of the membership matrix $\Pi$ can be obtained by decomposing the so-called symmetric Laplacian inverse matrix $\hat{M}$ via its leading $K$ eigenvectors. The symmetric Laplacian inverse matrix is defined as
\begin{align}
\hat{M}=\frac{\hat{W}+\hat{W}'}{2},
\end{align}
where $\hat{W}=(I-\alpha \hat{D}^{-1}A)^{-1}$, $I$ is an identity matrix, $\hat{D}$ is an $n\times n$ diagonal matrix such that its $i$-th diagonal entry is $\hat{D}(i,i)=\sum_{j=1}^{n}A(i,j)$, tuning parameters $\alpha=e^{-\gamma}$ and $\gamma>0$. As suggested by \cite{SLIM}, when forcing the diagonal elements of $\hat{M}$ to be 0, the performance is better. We also let $\hat{M}$'s diagonal entries be zeros in the numerical study.
\begin{remark}
This remark provides the motivation for using the SLIM matrix $\hat{M}$ to design spectral algorithms for community detection and the reason for $\alpha=e^{-\gamma}$. Since most real-world networks are sparse, $A$ has many zero entries. How to efficiently depict the closeness between nodes is challenging. \cite{SLIM} finds that using the first hitting time between two nodes is a good measure of closeness. Let $h_{ij}$ be the first hitting time from node $i$ to node $j$ to describe the closeness between nodes $i$ and $j$. Then, $\mathbb{E}(e^{-\gamma h_{ij}})$ is a good measure for nodes similarity, where the exponential transformation is used to down-weight the large first hitting time \cite{SLIM}. Since it is challenging to compute $\mathbb{E}(e^{-\gamma h_{ij}})$ directly, \cite{SLIM} approximates it by the $(i,j)$-th entry of the first hitting time matrix $H$, where $H$ is defined as $H=\sum_{k=1}^{\infty}e^{-\gamma k}(\hat{D}^{-1}A)^{k}=\sum_{k=1}^{\infty}\alpha^{k}(\hat{D}^{-1}A)^{k}=(I-\alpha \hat{D}^{-1}A)^{-1}-I$. Since $\hat{W}$ has the same eigenvectors as the first time hitting matrix $H$, $\hat{M}$ inherits the advantage of $H$ to depict the closeness between nodes. We see that $\alpha=e^{-\gamma}$ serves as a coefficient parameter to control the local similarity measure between nodes $i$ and $j$.
\end{remark}
We calculate the leading $K$ eigenvectors with a unit-norm of $\hat{M}$, and combine them to an $n\times K$ matrix:
\begin{equation*}
	\hat{X}=[\hat{\eta}_{1}, \hat{\eta}_{2},\ldots, \hat{\eta}_{K}].
\end{equation*}
Then we normalize each row of $\hat{X}$  to have unit length and denote the normalized matrix as $\hat{X}^{*}$,  i.e., the element of $i$-th row and $j$-th column of $\hat{X}^{*}$ is computed by
$$\hat{X}^{*}_{ij}=\hat{X}_{ij}/(\sum_{j=1}^{K}\hat{X}_{ij}^{2})^{1/2}, i = 1, \dots, n, j=1, \dots, K.$$

After obtaining the normalized matrix $\hat{X}^{*}$, we apply K-medians clustering on the rows of $\hat{X}^{*}$ to find cluster centers. The estimated cluster centers are $\hat{v}_{1}, \hat{v}_{2}, \ldots, \hat{v}_{K} \in\mathcal{R}^{1\times K}$,
\begin{align}
	\{\hat{v}_{1}, \ldots, \hat{v}_{K}\}=\mathrm{arg~}\underset{\hat{u}_{1},  \ldots, \hat{u}_{K}}{\mathrm{min}}\frac{1}{n}\sum_{i=1}^{n}\underset{\hat{u}\in\{\hat{u}_{1},\ldots, \hat{u}_{K}\}}{\mathrm{min}}\|\hat{X}^{*}_{i}-\hat{u}\|_{2}.
\end{align}
For convenience, denote $\hat{V}=(\hat{v}_{1}, \hat{v}_{2}, \ldots, \hat{v}_{K})'$, thus $\hat{V}$ is a $K\times K$ matrix with $i$-th row $\hat{v}_{i}, 1\leq i\leq K$.

Unlike classical spectral clustering methods for (non-mixed membership) community detection, we need a membership reconstruction step to obtain the final membership matrix. It can be produced as follows:
project the rows of $\hat{X}^{*}$ onto the spans of $\hat{v}_{1}, \ldots, \hat{v}_{K}$, i.e., compute the $n\times K$ projection matrix $\hat{Y}$ by \begin{equation}
	\hat{Y}=\hat{X}^{*}\hat{V}'(\hat{V}\hat{V}')^{-1}.
\end{equation}
It needs to be noted that there may exist some negative entries of $\hat{X}^{*}\hat{V}'(\hat{V}\hat{V}')^{-1}$, thus we set $\hat{Y} = \mathrm{max}(0, \hat{Y})$.
Then we can estimate $\pi_{i}$ by
\begin{equation}\label{pi}
	\hat{\pi}_{i}=\hat{Y}_{i}/\|\hat{Y}_{i}\|_{1}, 1\leq i\leq n.
\end{equation}
Finally, the estimated membership matrix $\hat{\Pi}$ is \begin{equation}
	\hat{\Pi}=(\hat{\pi}_{1},\hat{\pi}_{2}, ...,\hat{\pi}_{n})'.
\end{equation}
 If all the $K$ entries of $\hat{Y}_{i}$ (the $i$-th row of $\hat{Y}$) are negative, we set $\hat{Y}_{i}=-\hat{Y}_{i}$ to avoid the case that $\|\hat{Y}_{i}\|_{1}=0$ for any $i$.
\begin{remark}
Similar as \cite{SLIM}, we can also replace $A$ and $\hat{D}$ in matrix $\hat{W}$ with $A_{\tau}=A+\tau I$ and $\hat{D}_{\tau}=\mathrm{diag}(A_{\tau}\times 1_{n\times 1})$. We call the method with this regularization by Mixed-$\mathrm{SLIM}_{\tau}$. For simplicity, we use `Mixed-SLIM methods' to denote all methods that apply the SLIM matrix for mixed membership community detection problems. Note that as studied in \cite{amini2013pseudo,RSC,joseph2016impact,SLIM}, a positive regularizer $\tau$ is applied to make $\hat{D}^{-1}_{\tau}$ be stable since most real-world networks are sparse and $A$ contains many zero elements.
\end{remark}
\begin{remark}
As suggested by \cite{SLIM}, when handing large networks in practice, we suggest to approximate $\hat{W}$ by $\sum_{t=1}^{T}\alpha^{t}(\hat{D}^{-1}A)^{t}$, instead of calculating the inverse of $I-\alpha \hat{D}^{-1}A$ directly. This approximation approach will be examined with empirical datasets in Section \ref{sec5} where it is referred to as Mixed-$\mathrm{SLIM}_{appro}$ and Mixed-$\mathrm{SLIM}_{\tau appro}$ for the regularized version.
\end{remark}
\subsection{Choice of tuning parameters}
There are three tuning parameters, $\tau$,$\gamma$ and $T$ to be chosen in Mixed-SLIM methods.

The choice of $\tau$ is flexible. We can set $\tau=c\bar{d}$ where $c$ is a positive constant, the average degree $\bar{d}$ is computed by $\bar{d}=\sum_{i=1}^{n}\hat{D}(i,i)/n$, or set it as $\tau=cd_{\mathrm{max}}$ where $d_{max}=\mathrm{max}_{1\leq i\leq n}\hat{D}(i,i)$, and we can also set it as $\tau=c\frac{d_{\mathrm{max}}+d_{\mathrm{min}}}{2}$  where $d_{min}=\mathrm{min}_{1\leq i\leq n}\hat{D}(i,i)$.  Empirically, we set $\tau=0.1\bar{d}$ in simulation and empirical studies. Same as \cite{SLIM}, a good choice of $\gamma$ is 0.25 which provides satisfactory performances for Mixed-SLIM methods. For Mixed-$\mathrm{SLIM}_{appro}$ and Mixed-$\mathrm{SLIM}_{\tau appro}$, we set the default value for $T$ as 10 in this article and we also study the	insensitivity on $T$ for Mixed-$\mathrm{SLIM}_{appro}$ in Section \ref{StudyT}.
\section{Theoretical results}\label{sec4}
In this section, we show the asymptotic consistency of Mixed-$\mathrm{SLIM}_{\tau}$ under the DCMM model. The consistency of Mixed-SLIM can be produced similarly.  For the convenience of demonstrating the theoretical results, we first define items given $\Omega$ under DCMM. Set $D_{\tau}=D+\tau I, \Omega_{\tau}=\Omega+\tau I$. Set $W_{\tau}=(I-\alpha D_{\tau}^{-1}\Omega_{\tau})^{-1}$. Set $M_{\tau}=\frac{W_{\tau}+W_{\tau}'}{2}$ and force the diagonal entries of $M_{\tau}$ to 0.  Let $X_{\tau}$ be the $n\times K$ matrix such that $X_{\tau}=[\eta_{1}, \eta_{2},\ldots, \eta_{K}]$, where $\{\eta_{k}\}^{K}_{k=1}$ are the leading $K$ eigenvectors with unit-norm of $M_{\tau}$. Let $X_{\tau}^{*}$ be the matrix obtained by normalizing each row of $X_{\tau}$  to have unit length. Let $v_{1}, v_{2}, \ldots, v_{K} \in\mathcal{R}^{1\times K}$ be the $K$ estimated cluster centers obtained by applying K-medians on matrix $X_{\tau}^{*}$. Let $V_{\tau}$ be a $K\times K$ whose $k$-th row is $v_{k}, 1\leq k\leq K$. Let $Y_{\tau}$ be the $n\times K$ matrix such that $Y_{\tau}=X_{\tau}^{*}V_{\tau}'(V_{\tau}V_{\tau}')^{-1}$.

Set $\theta_{\mathrm{max}}=\mathrm{max}_{i}\theta(i)$. Throughout this paper, for our theoretical analysis, $\alpha$ is assumed to be a small positive value. To control network sparsity for our theoretical analysis,  we need the following assumption
\begin{assumption}\label{A1}
Assume $\theta_{\mathrm{max}}\|\theta\|_{1}\geq \mathrm{log}(n)$.
\end{assumption}
We use subscript $\tau$ to tab terms obtained from  Mixed-$\mathrm{SLIM}_{\tau}$ algorithm. By considering this notation, we can have  $\hat{W}_{\tau}, \hat{M}_{\tau}, \hat{X}_{\tau}, \hat{X}^{*}_{\tau}, \hat{V}_{\tau}, \hat{Y}_{\tau}$.  The following lemma bounds $\|\hat{M}_{\tau}-M_{\tau}\|$.
\begin{lemma}\label{boundM}
Under $DCMM(n, P, \Theta, \Pi)$, if assumption \ref{A1} holds, with probability at least $1-o(n^{-3})$,
\begin{align*}
\|\hat{M}_{\tau}-M_{\tau}\|=O(\frac{\alpha\sqrt{\mathrm{log}(n)\theta_{\mathrm{max}}\|\theta\|_{1}}}{\tau+\alpha^2}).
\end{align*}
\end{lemma}
Set $err_{n}=\|\hat{M}_{\tau}-M_{\tau}\|$ for convenience. Similar as in \cite{OCCAM}, we define the Hausdorff distance which is used to measure the dissimilarity between two cluster centers as $D_{H}(S,T)=\mathrm{min}_{\sigma\in \Sigma}\|S-T\sigma\|_{F}$ for any $K\times K$ matrix $S$ and $T$ where $\Sigma$ is the set of $K\times K$ permutation matrix. The sample loss function for K-medians is defined by
\begin{align*}
\mathcal{L}_{n}(Q,S)=\frac{1}{n}\sum_{i=1}^{n}\mathrm{min}_{1\leq k\leq K}\|Q_{i}-S_{k}\|_{F},
\end{align*}
where $Q\in \mathcal{R}^{n\times K}$ is a matrix whose rows $Q_{i}$ are vectors to be clustered, and $S\in \mathcal{R}^{K\times K}$ is a matrix whose rows $S_{k}$ are cluster centers. Assuming the rows of $Q$ are i.i.d. random vectors sampled from a distribution $\mathcal{G}$, we similarly define the population loss function for K-medians by
\begin{align*}
\mathcal{L}(\mathcal{G};S)=\int \mathrm{min}_{1\leq k\leq K}\|x-S_{k}\|_{F}d\mathcal{G}.
\end{align*}
Let $\mathcal{F}_{\tau}$ be the distribution of $(X_{\tau}^{*})_{i}$,  assume the following condition on $\mathcal{F}_{\tau}$ holds:
\begin{assumption}\label{A2}
Let
$V_{\mathcal{F}_{\tau}}=\mathrm{arg~min}_{U}\mathcal{L}(\mathcal{F}_{\tau};U)$ be the global minimizer of the population loss function $\mathcal{L}(\mathcal{F}_{\tau};U).$ Then $V_{\mathcal{F}_{\tau}}=V_{\tau}$ up to a row permutation. Further, there
exists a global constant $\kappa$ such that $\kappa K^{-1}D_{H}(U,V_{\mathcal{F}_{\tau}})\leq\mathcal{L}(\mathcal{F}_{\tau};U)-\mathcal{L}(\mathcal{F}_{\tau};V_{\mathcal{F}_{\tau}})$ for all $U$.
\end{assumption}
Assumption \ref{A2} essentially states that the population K-medians loss function, which is determined by $\mathcal{F}_{\tau}$, has a unique minimum at the right place.
Then we have the following lemma.
\begin{lemma}\label{boundV}
Under $DCMM(n,P,\Theta,\Pi)$, if assumptions \ref{A1} and \ref{A2} hold, with probability at least $\mathrm{Pr}(n, K)$, there exists an orthogonal matrix $\hat{O}$ such that
\begin{align*}
\|\hat{V}_{\tau}\hat{O}-V_{\tau}\|_{F}=O(\frac{K^{1.5}err_{n}}{m\sqrt{n}|\lambda_{K}-\lambda_{K+1}|}),
\end{align*}
where $m=\mathrm{min}_{i}\{\mathrm{min}\{\|(\hat{X}_{\tau})_{i}\|, \|(X_{\tau})_{i}\|\}\}$ is the length of the shortest row in $\hat{X}_{\tau}$ and $X_{\tau}$,  $\mathrm{Pr}(n,K)\rightarrow 1$ as $n\rightarrow \infty$.
\end{lemma}
To bound $\|\hat{Y}_{\tau}-Y_{\tau}\|_{F}$, we need the following assumption.
\begin{assumption}\label{A3}
There exists a global constant $m_{V_{\tau}}>0$ such that $\lambda_{\mathrm{min}}(V_{\tau}V_{\tau}')\geq m_{V_{\tau}}$
\end{assumption}
\begin{lemma}\label{boundY}
Under $DCMM(n,P,\Theta,\Pi)$, if assumptions \ref{A1}, \ref{A2} and \ref{A3} hold, with probability at least $\mathrm{Pr}(n,K)$,
\begin{align*}
\|\hat{Y}_{\tau}-Y_{\tau}\|_{F}=O(\frac{K^{1.5}err_{n}}{m|\lambda_{K}-\lambda_{K+1}|m_{V_{\tau}}}).
\end{align*}
\end{lemma}
Lemma \ref{boundY} is helpful to obtain the bound of the Mixed-Hamming error rate of Mixed-SLIM since we obtain $\hat{\Pi}$ from $\hat{Y}_{\tau}$ directly. The next theorem is the main theoretical result of this paper, which provides a theoretical bound on $\|\hat{\Pi}-\Pi\|_{F}$.

\begin{theorem}\label{main}
Under $DCMM(n,P,\Theta,\Pi)$, set $m_{Y_{\tau}}=\mathrm{min}_{i}\{\|(Y_{\tau})_{i}\|\}$ as the shortest length of rows in $Y_{\tau}$. If assumptions \ref{A1}--\ref{A3} hold, then with probability at least $\mathrm{Pr}(n,K)$, we have
\begin{align*}
\frac{\|\hat{\Pi}-\Pi\|_{F}}{\sqrt{n}}=O(\frac{K^{1.5}err_{n}}{m|\lambda_{K}-\lambda_{K+1}|m_{V_{\tau}}m_{Y_{\tau}}\sqrt{n}}).
\end{align*}
\end{theorem}
\section{Numerical Results}\label{sec5}
In this section, first, we investigate the performances of Mixed-SLIM methods for the problem of mixed membership community detection via synthetic data. Then we apply some real-world networks with true label information to test Mixed-SLIM methods' performances for community detection, and we apply the SNAP ego-networks \citep{leskovec2012learning} to investigate their performances for mixed membership community detection. We omit the numerical results on synthetic data experiments for Mixed-$\mathrm{SLIM}_{\tau}$, Mixed-$\mathrm{SLIM}_{appro}$ and Mixed-$\mathrm{SLIM}_{\tau appro}$ since they behave similarly to Mixed-SLIM. Finally, we provide a discussion on the choice of $T$ for Mixed-$\mathrm{SLIM}_{appro}$.
\subsection{Synthetic data experiments}
To measure the performance for any approach focusing on mixed membership community detection, we apply the mixed-Hamming error rate defined below:
\begin{align*}
\mathrm{min}_{O\in\{ K\times K\mathrm{permutation~matrix}\}}\frac{1}{n}\|\hat{\Pi}O-\Pi\|_{1},
\end{align*}
where $\Pi$ and $\hat{\Pi}$ are the true and estimated mixed membership matrices respectively. For simplicity, we write the mixed-Hamming error rate as $\sum_{i=1}^{n}\|\hat{\pi}_{i}-\pi_{i}\|_{1}/n$. For all the experiments in this section, we always report the mean of the mixed-Hamming error rates for all approaches. Therefore for all the figures in this section, the y-axis always records the mean of $\sum_{i=1}^{n}\|\hat{\pi}_{i}-\pi_{i}\|_{1}/n$ over 50 repetitions.

We investigate the performances of our  Mixed-SLIM methods by comparing them with Mixed-SCORE \citep{mixedSCORE}, OCCAM \citep{OCCAM} and GeoNMF \citep{GeoNMF} via Experiment 1 which contains 12 sub-experiments.

\textbf{\texttt{Experiment 1}}. Unless specified, we set $n=500$ and $K=3$. For $0\leq n_{0}\leq 160$, let each block own $n_{0}$ number of pure nodes. For the top $3n_{0}$ nodes $\{1,2, \ldots, 3n_{0}\}$, we let these nodes be pure and let nodes $\{3n_{0}+1, 3n_{0}+2,\ldots, 500\}$ be mixed. Fixing $x\in [0, \dfrac{1}{2})$, let all the mixed nodes have four different memberships $(x, x, 1-2x), (x, 1-2x, x), (1-2x, x, x)$ and $(1/3,1/3,1/3)$, each with $\dfrac{500-3n_{0}}{4}$ number of nodes. Fixing $\rho\in(0, 1)$, the mixing matrix $P$ has diagonals 0.5 and off-diagonals $\rho$. In our experiments, unless specified, there are two settings about $\theta$, one is $\theta(i)=0.4$ (i.e., the MMSB case), another is $\theta(i)=0.2+0.8(i/n)^{2}$ (i.e., the DCMM case). The details of the experiments are described as follows.
\subsubsection{Fraction of pure nodes}
In experiments 1(a) and 1(b), we study how the fraction of pure nodes affects the behaviors of these mixed membership community detection methods under MMSB and DCMM, respectively. We fix $(x,\rho)=(0.4, 0.1)$ and let $n_{0}$ range in $\{40, 60, 80, 100, 120, 140, 160\}$. In \textit{Experiment 1(a)} generate $\theta$ as $\theta(i)=0.4$ for all $1\leq i\leq n$, that is, it is under MMSB model. In \textit{Experiment 1(b)}, generate $\theta$ as $\theta(i)=0.2+0.8(i/n)^{2}$ for all $1\leq i\leq n$, i.e., it is under DCMM model.

Numerical results of these two sub-experiments are shown in panels (a) and (b) of Figure \ref{EX1}, respectively. From the results in subfigure 1(a), it can be found that Mixed-SLIM performs similarly to Mixed-SCORE while both two methods perform better than OCCAM and GeoNMF under the MMSB setting. Subfigure 1(b) suggests that Mixed-SLIM significantly outperforms Mixed-SCORE, OCCAM, and GeoNMF under the DCMM setting. It is interesting to find that only Mixed-SLIM enjoys better performances as the fraction of pure nodes increases under the DCMM setting.
\subsubsection{Connectivity across communities}
In experiments 1(c) and 1(d), we study how the connectivity (i.e., $\rho$, the off-diagonal entries of $P$) across communities under different settings affects the performances of these methods. Fix $(x,n_{0})=(0.4, 100)$ and let $\rho$ range in $\{0, 0.02, 0.04, \ldots, 0.2\}$.  In \textit{Experiment 1(c)}, $\theta$ is  generated from MMSB model. In \textit{Experiment 1(d)},  $\theta$ is  generated from DCMM model.

Numerical results of these two sub-experiments are shown in panels (c) and (d) of Figure \ref{EX1}. From subfigure (c), under the MMSB model, we can find that  Mixed-SLIM, Mixed-SCORE, OCCAM, and GeoNMF have similar performances, and as $\rho$ increases they all perform poorer. Under the DCMM model, the mixed Humming error rate of Mixed-SLIM decreases as $\rho$ decreases, while the performances of the other three approaches are still unsatisfactory.
\subsubsection{Purity of mixed nodes}
We study how the purity of mixed nodes under different settings affects the performances of these overlapping community detection methods in sub-experiments 1(e) and 1(f). Fix $(n_{0},\rho)=(100, 0.1)$, and let $x$ range in $\{0, 0.05, \ldots, 0.5\}$. In \textit{Experiment 1(e)}, $\theta$ is  generated from MMSB model. In \textit{Experiment 1(f)},  $\theta$ is  generated from DCMM model.

Panels (e) and (f) of Figure \ref{EX1} report the numerical results of these two sub-experiments. They suggest that estimating the memberships becomes harder as the purity of mixed nodes decreases. Mixed-SLIM and Mixed-SCORE perform similarly and both two approaches perform better than OCCAM and GeoNMF under the MMSB setting. Meanwhile, Mixed-SLIM significantly outperforms the other three methods under the DCMM setting.
\subsubsection{Degree heterogeneity}
We study how degree heterogeneity affects the performances of these overlapping community detection methods here.  Fix $(n_{0},\rho, x)=(100, 0.1, 0.4)$.

\textit{Experiment 1(g)}: Let $z_{1}$ range in $\{1,2,\ldots, 8\}$. Generate $\theta$ as $1/\theta(i)\overset{\mathrm{iid}}{\sim} U(1,z_{1})$ for $i=1,2,\ldots,n$  where $U(1,z_{1})$ denotes the uniform distribution on $[1,z_{1}]$. Note that a larger $z_{1}$ implies more heterogeneous and therefore it is more challenging to detect communities.

\textit{Experiment 1(h)}: Let $z_{2}$ range in $\{0, 0.1,\ldots, 0.8\}$. Generate $\theta$ as $\theta(i)=0.1+z_{2}+0.4(i/n)^{2}$ for $i=1,2,\ldots,n$.

Numerical results of these two sub-experiments are shown in panels (g) and (h) of Figure \ref{EX1}, from which we can find that Mixed-SLIM outperforms the other three approaches in these two settings.
\subsubsection{Sparsity of mixing matrix}
We study how the sparsity of the mixing matrix affects the behaviors of these mixed membership community detection methods in sub-experiments 1(i) and 1(j).
Fix $(x, n_{0})=(0.4, 100)$, let $p\in\{1,1.5,\ldots,5\}$. Set $P_{(ij)}$ as
\[\renewcommand{\arraystretch}{0.75}
P_{(ij)}
=
p\begin{bmatrix}
   0.2 & 0.05 &0.05\\
   0.05& 0.2  &0.05\\
   0.05& 0.05 &0.2\\
\end{bmatrix}.
\]
Therefore, a larger $p$ indicates a denser simulated network.
In \textit{Experiment 1(i)}, $\theta$ is  generated from MMSB model. In \textit{Experiment 1(j)},  $\theta$ is  generated from DCMM model.

\begin{figure}
\centering
\subfigure[Experiment 1(a)]{
\includegraphics[width=0.2\textwidth]{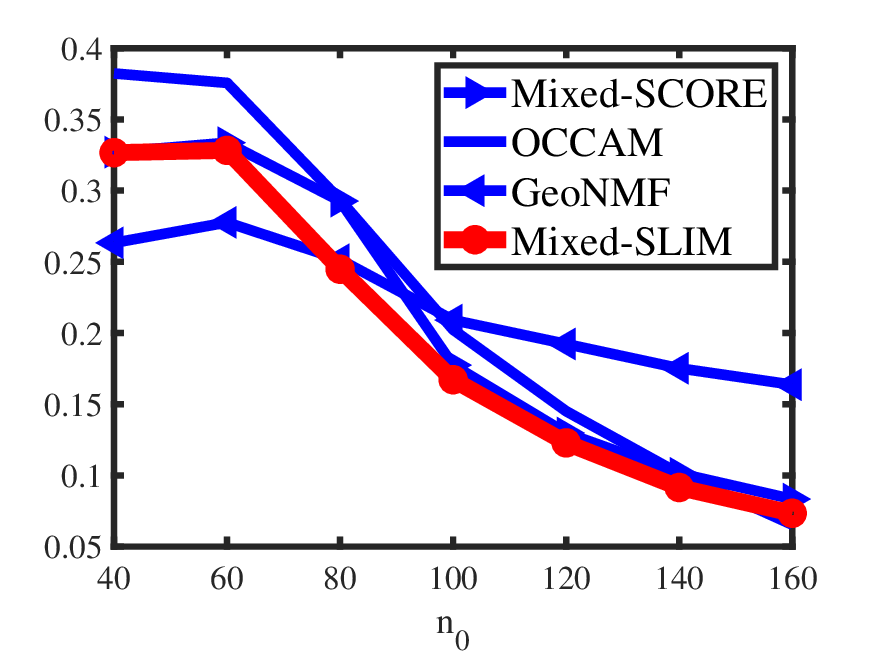}}
\subfigure[Experiment 1(b)]{
\includegraphics[width=0.2\textwidth]{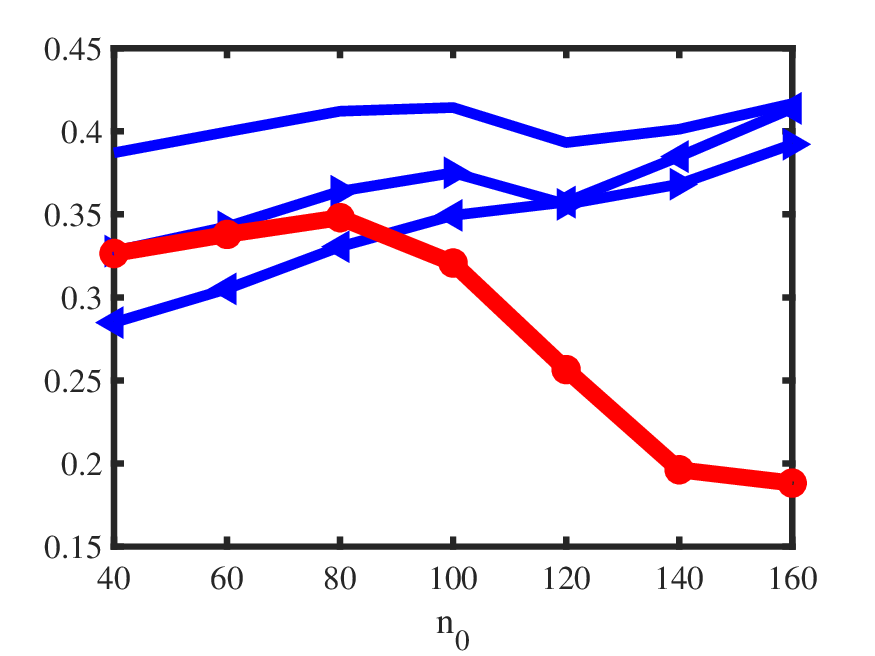}}
\subfigure[Experiment 1(c)]{
\includegraphics[width=0.2\textwidth]{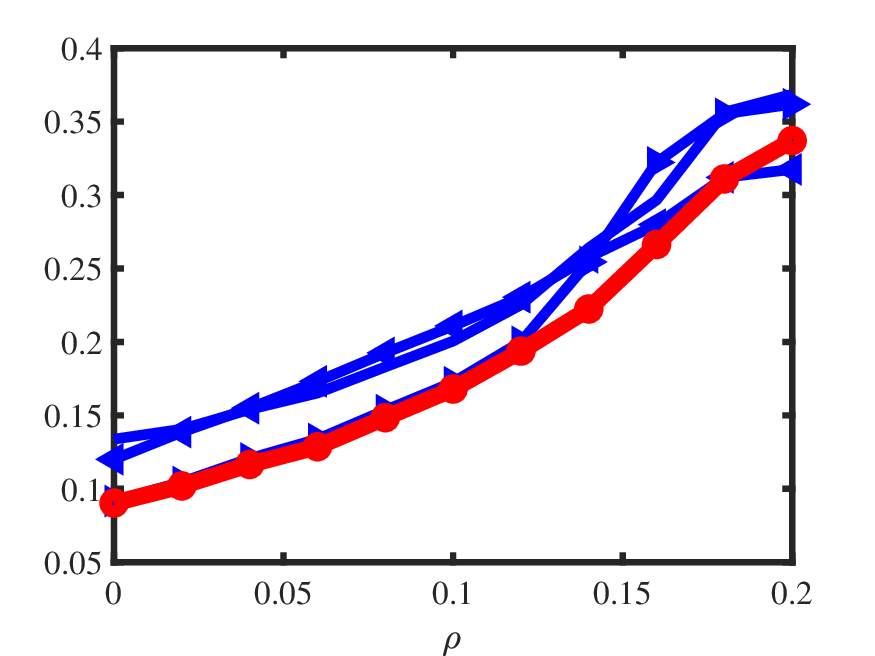}}
\subfigure[Experiment 1(d)]{
\includegraphics[width=0.2\textwidth]{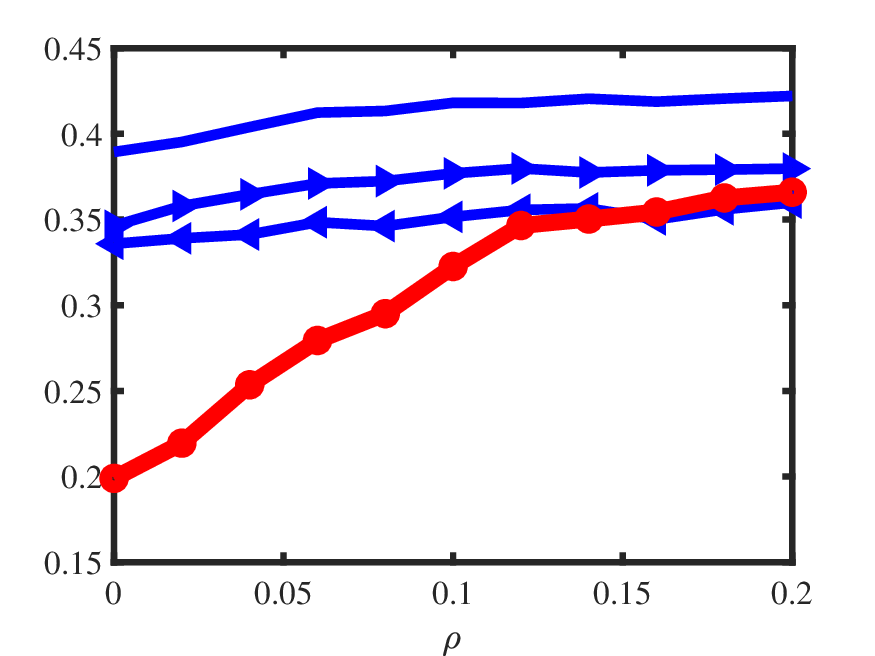}}
\subfigure[Experiment 1(e)]{
\includegraphics[width=0.2\textwidth]{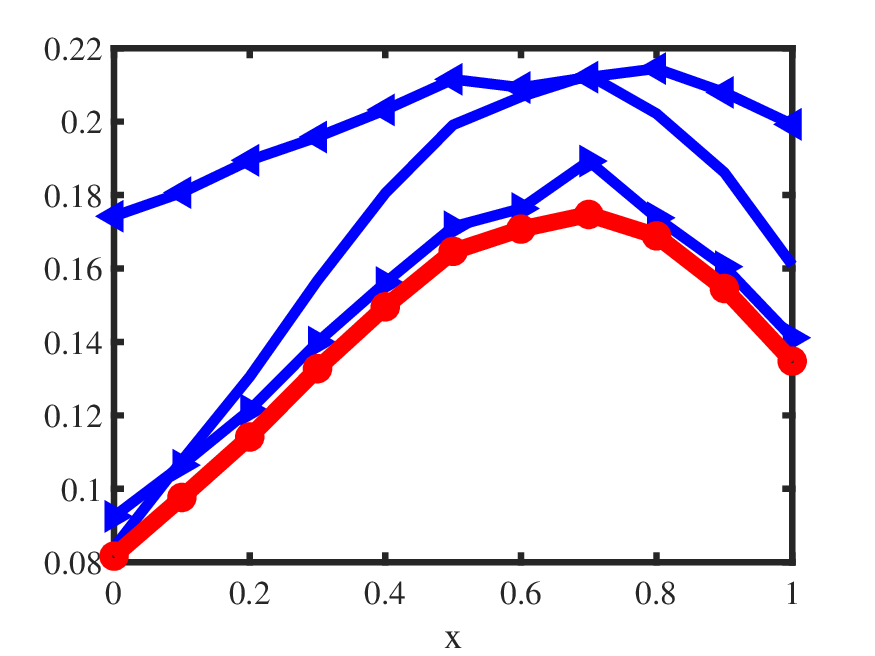}}
\subfigure[Experiment 1(f)]{
\includegraphics[width=0.2\textwidth]{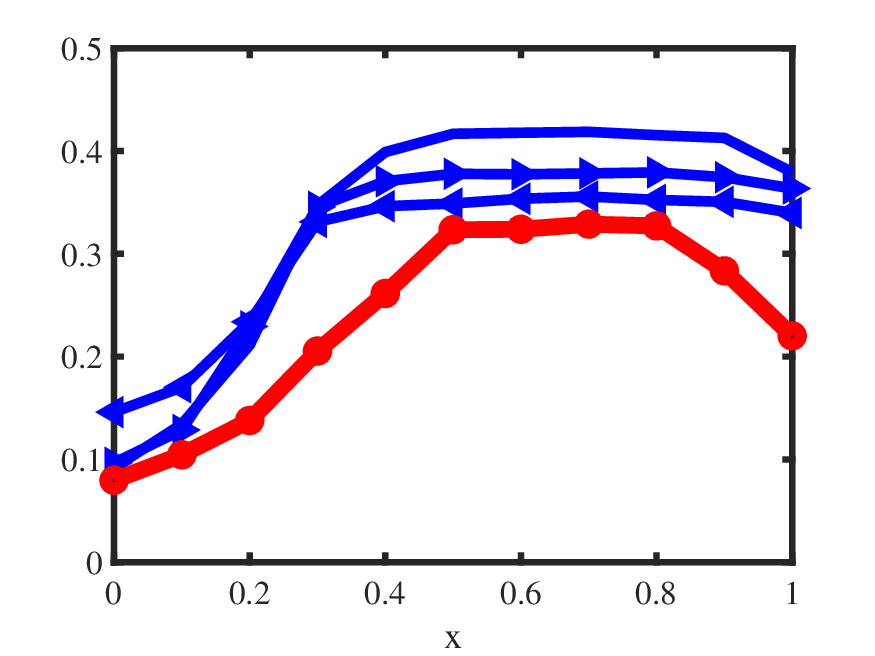}}
\subfigure[Experiment 1(g)]{
\includegraphics[width=0.2\textwidth]{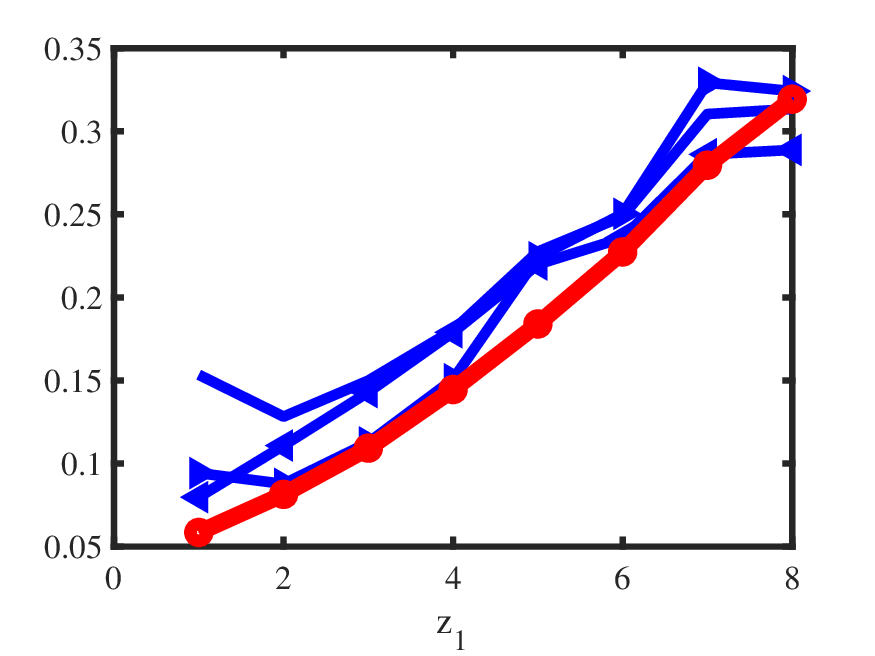}}
\subfigure[Experiment 1(h)]{
\includegraphics[width=0.2\textwidth]{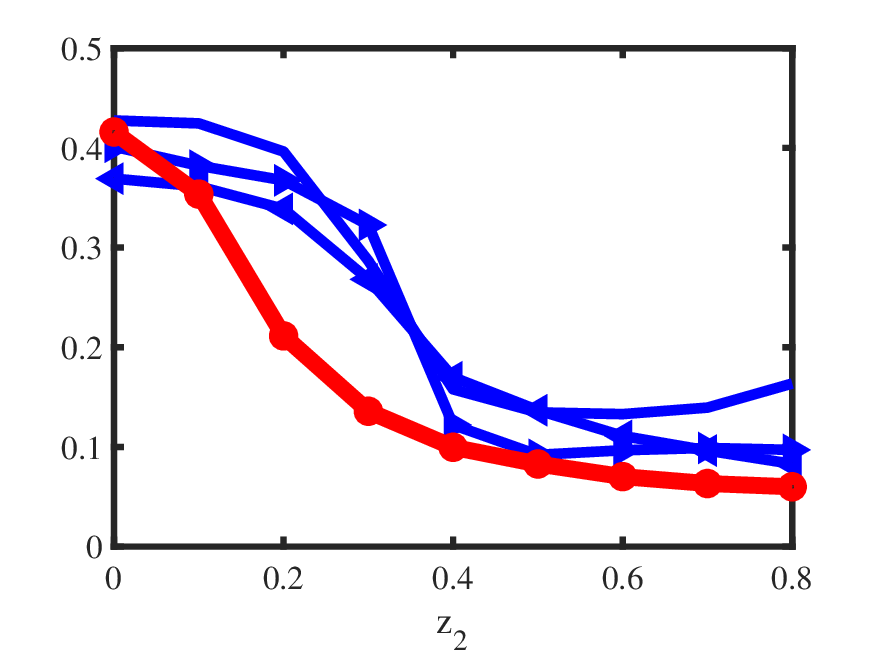}}
\subfigure[Experiment 1(i)]{
\includegraphics[width=0.2\textwidth]{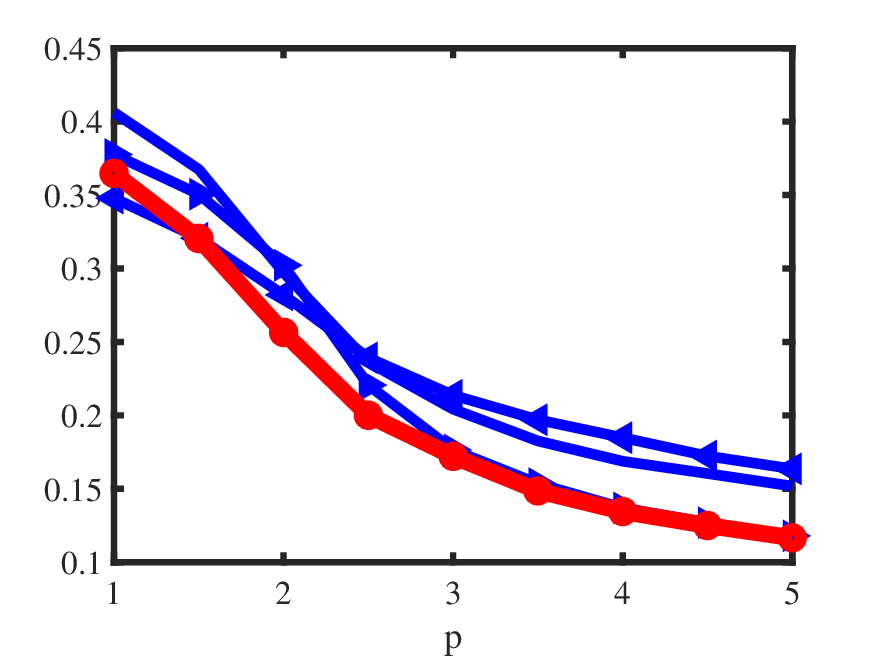}}
\subfigure[Experiment 1(j)]{
\includegraphics[width=0.2\textwidth]{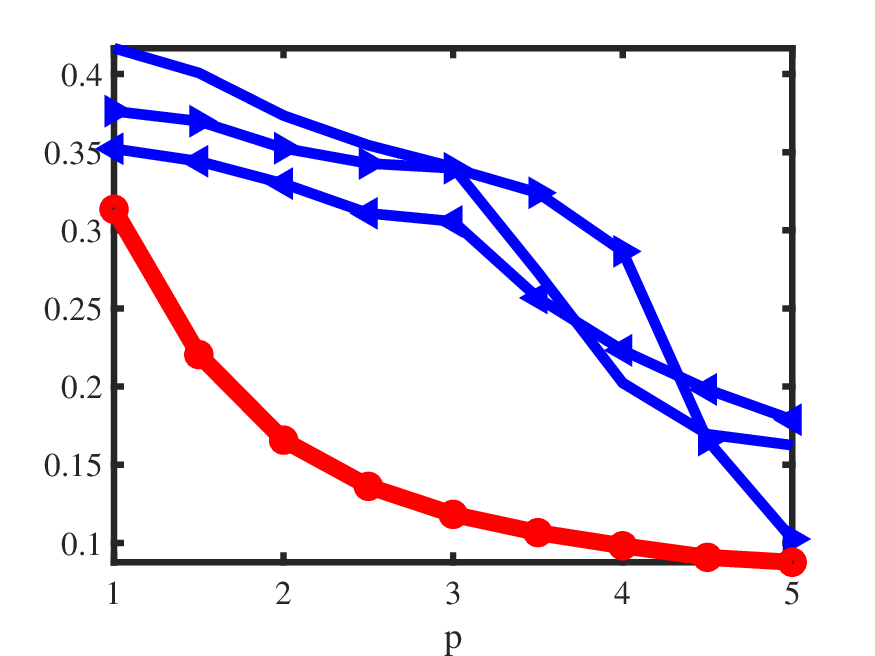}}
\subfigure[Experiment 1(k)]{
\includegraphics[width=0.2\textwidth]{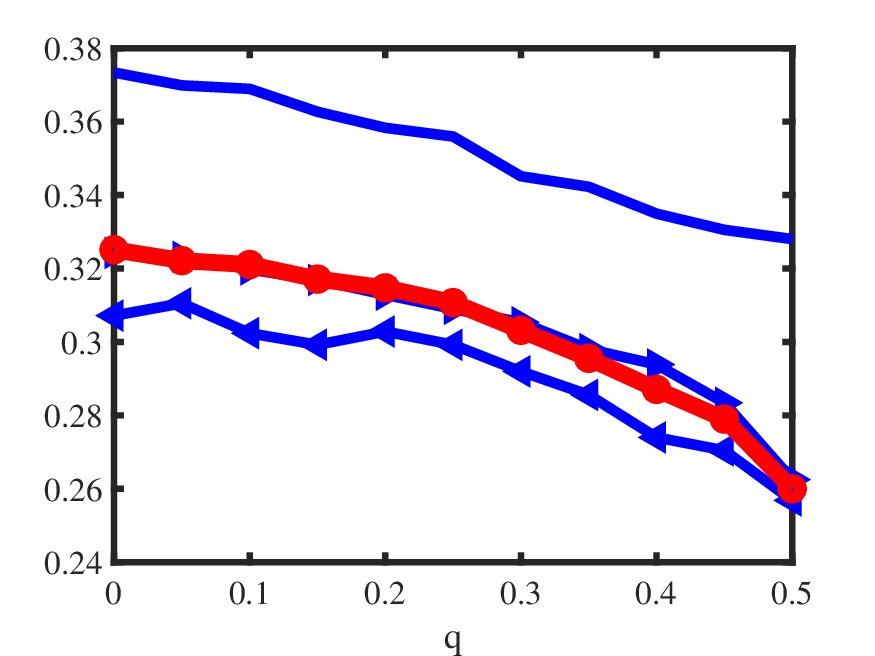}}
\subfigure[Experiment 1(l)]{
\includegraphics[width=0.2\textwidth]{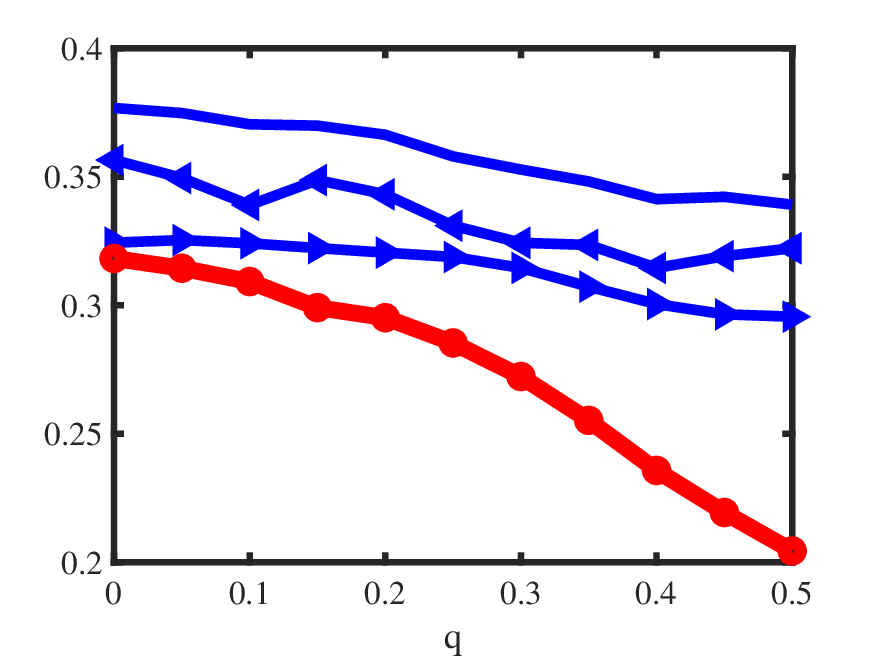}}
\caption{Estimation errors of Experiment 1 (y-axis: $\sum_{i=1}^{n}n^{-1}\|\hat{\pi}_{i}-\pi_{i}\|_{1}$).}
\label{EX1} 
\vskip -0.2in
\end{figure}

The numerical results of these two sub-experiments are shown in panels (i) and (j) of Figure \ref{EX1}, from which we can find that: all procedures enjoy improvement performances when the simulated network becomes denser; Mixed-SLIM outperforms the other three approaches, especially under the DCMM setting.
\subsubsection{Detectability of clusters}
We study how the detectability of clusters affects the behaviors of these mixed membership community detection methods in sub-experiments 1(k) and 1(l). In Experiment 1(k) and 1(l), set $(n, K, n_{0}, x)=(500, 4, 75, 0.2)$. For the top $4n_{0}$ nodes $\{1,2,\ldots, 4n_{0}\}$, let these nodes be pure and let nodes $\{4n_{0}+1, 4n_{0}+2,\ldots, 500\}$ be mixed. Let all the mixed nodes have five different memberships $(x,x,x,1-3x), (x,x,1-3x,x), (x,1-3x,x,x),$ $(1-3x,x,x,x), (1/4,1/4,1/4,1/4)$. Let $q$ range in $\{0, 0.05, 0.1, \ldots, 0.5\}$.
The mixing matrix $P_{(kl)}$ is set as
\[\renewcommand{\arraystretch}{0.75}
P_{(kl)}
=
\begin{bmatrix}
    0.5+q&0.5&0.3&0.3\\
    0.5&0.5+q&0.3&0.3\\
    0.3&0.3&0.5+q&0.5\\
    0.3&0.3&0.5&0.5+q\\
\end{bmatrix}.
\]
When $q=0$, the networks have only two communities, and as $q$ increases, the four communities become more distinguishable. \textit{Experiment 1(k)} is the case under MMSB model, and \textit{Experiment 1(l)} is under DCMM model.

The numerical results are given by the last two panels of Figure \ref{EX1}. Subfigure 1(k) suggests that Mixed-SLIM, Mixed-SCORE, and GeoNMF share similar performances and they perform better than OCCAM under the MMSB setting. the proposed Mixed-SLIM significantly outperforms the other three methods under the DCMM setting.
\subsection{Application to real-world datasets for community detection}\label{secreal8}
In this section, four real-world network datasets with known label information are analyzed to test the performances of our Mixed-SLIM methods for community detection. The four datasets can be downloaded from
\url{ http://www-personal.umich.edu/ ~ mejn/netdata/}. For the four datasets, the true labels are suggested by the original authors, and they are regarded as the ``ground truth'' to investigate the performances of Mixed-SLIM methods in this paper.
A brief introduction of the four datasets is given below.
\begin{itemize}
\item \textbf{Dolphins}: this network consists of frequent associations between 62 dolphins in a community living off Doubtful Sound. In the Dolphins network, node denotes a dolphin, and edge stands for companionship \cite{dolphins0, dolphins1, dolphins2}. The network splits naturally into two large groups females and males \cite{dolphins1, dolphinnewman}, which are seen as the ground truth in our analysis.
\item \textbf{Polbooks}: this network is about US politics
    published around the 2004 presidential election and sold by the online bookseller Amazon.com. In Polbooks, nodes represent books, edges represent frequent co-purchasing of books by the same buyers. Full information about edges and labels can be downloaded from \url{http://www-personal.umich.edu/~mejn/netdata/}. The original network contains 105 nodes labeled as either ``Conservative”, ``Liberal”, or ``Neutral”. Nodes labeled ``Neutral'' are removed for community detection in this paper.
\item \textbf{UKfaculty}: this network reflects the friendship among academic staff of a given Faculty in a UK university consisting of three separate schools \cite{UKfaculty}. The original network contains 81 nodes, and the smallest group only has 2 nodes. The smallest group is removed for community detection in this paper.
\item \textbf{Polblogs}: this network consists of political blogs during the
	2004 US presidential election \citep{Polblogs1}. Each blog belongs to one of the two parties liberal or
	conservative. As suggested by \cite{DCSBM}, we only consider the largest connected component with 1222 nodes and ignore the edge direction for community detection.
\end{itemize}
Before comparing these methods, we take some preprocessing to remove nodes that may have mixed memberships for community detection. For the Polbooks data, nodes labeled as ``neutral'' are removed. The smallest group with only 2 nodes in UKfaculty data is removed. Table \ref{real4} presents some basic information about the four datasets.
\begin{table}[h!]
\centering
\caption{Four real-world data sets with known label information.}
\label{real4}
\begin{tabular}{cccccccccc}
\toprule
\#&Dolphins&Polbooks&UKfaculty&Polblogs\\
\midrule
$n$&62&92&79&1222\\
$K$&2&2&3&2\\
$d_{\mathrm{min}}$&1&1&2&1\\
$d_{\mathrm{max}}$&12&24&39&351\\
\bottomrule
\end{tabular}
\end{table}
\begin{table}[h!]
\footnotesize
\centering
\caption{Error rates on the four empirical data sets.}
\label{real4errors}
\resizebox{\columnwidth}{!}{
\begin{tabular}{cccccccccc}
\toprule
\textbf{ Methods} &Dolphins&Polbooks&UKfaculty&Polblogs\\
\midrule
SCORE&\textbf{0/62}&\textbf{1/92}&1/79&58/1222\\
SLIM&\textbf{0/62}&2/92&1/79&51/1222\\
OCCAM&1/62&3/92&5/79&60/1222\\
Mixed-SCORE&2/62&3/92&6/79&60/1222\\
GeoNMF&1/62&3/92&4/79&64/1222\\
\hline
Mixed-SLIM&\textbf{0/62}&2/92&\textbf{0/79}&\textbf{49/1222}\\
Mixed-$\mathrm{SLIM}_{\tau}$&\textbf{0/62}&2/92&\textbf{0/79}&51/1222\\
Mixed-$\mathrm{SLIM}_{appro}$&\textbf{0/62}&2/92&\textbf{0/79}&50/1222\\
Mixed-$\mathrm{SLIM}_{\tau appro}$&\textbf{0/62}&2/92&\textbf{0/79}&51/1222\\
\bottomrule
\end{tabular}}
\end{table}

Table \ref{real4errors} records the error rates on the four real-world networks. The numerical results suggest that Mixed-SLIM methods enjoy satisfactory performances compared with SCORE, SLIM, OCCAM, Mixed-SCORE, and GeoNMF when detecting the four empirical datasets. Especially, the number error for Mixed-SLIM on the Polblogs network is 49, which is the smallest number error for this dataset in literature as far as we know.
\subsection{Application to SNAP ego-networks for mixed membership community detection}
The ego-networks dataset contains more than 1000 ego-networks from Facebook, Twitter, and GooglePlus. In an ego-network, all the nodes are friends of one central user and the friendship groups or circles (depending on the platform) set by this user can be used as ground truth communities. The SNAP ego-networks are open to the public, and it can be downloaded from \url{ http://snap.stanford.edu/data/}. It is applied to test the performances of OCCAM \citep{OCCAM} after some preprocessing. We obtain the SNAP ego-networks parsed by Yuan Zhang (the first author of the OCCAM method \citep{OCCAM}). The parsed SNAP ego-networks are slightly different from those used in \cite{OCCAM}. To get a better sense of what the different social networks look like and how different characteristics potentially affect the performance of our Mixed-SLIM, we report the following summary statistics for each network: number of nodes $n$, number of communities $K$, and the proportion of overlapping nodes $r_{o}$, i.e., $r_{o}=\frac{\mathrm{number~of~nodes~with~mixed~membership}}{n}$. We report the means and standard deviations of these measures for each of the social networks in Table \ref{dataSNAP}.
\begin{table}[h!]
\centering
\caption{Mean (SD) of summary statistics for ego-networks.}
\label{dataSNAP}
\begin{tabular}{cccccccccc}
\toprule
&\#Networks&$n$&$K$&$r_{o}$\\
\midrule
Facebook&7&236.57&3& 0.0901\\
&-&(228.53)&(1.15)&(0.1118)\\
\hline
GooglePlus&58&433.22&2.22&0.0713 \\
&-&(327.70)&(0.46)&(0.0913)\\
\hline
Twitter&255&60.64&2.63&0.0865\\
&-&(30.77)&(0.83)&(0.1185)\\
\bottomrule
\end{tabular}
\end{table}

From Table \ref{dataSNAP}, we see that Facebook and GooglePlus networks tend to be larger than Twitter networks. Meanwhile, the proportions of overlapping nodes in Twitter networks tend to be larger than that of Facebook and GooglePlus networks.

We report the averaged mixed Hamming error rates for our methods and the other three competitors in Table \ref{ErrorSNAP}. Mixed-$\mathrm{SLIM}_{\tau appro}$ outperforms the other three Mixed-SLIM methods on all SNAP ego-networks and it significantly outperforms Mixed-SCORE, OCCAM, and GeoNMF on GooglePlus and Twitter networks.  Mixed-SLIM methods have smaller averaged mixed Humming error rates than Mixed-SCORE, OCCAM, and GeoNMF on the GooglePlus networks and Twitter networks, while they perform slightly poorer than Mixed-SCORE on Facebook networks. Meanwhile, we also find that OCCAM and GeoNMF share similar performances on the ego-networks. It is interesting to find that the error rates on Twitter and GooglePlus networks are higher than error rates on Facebook which may be because Twitter and GooglePlus networks have a higher proportion of overlapping nodes than Facebook.
\begin{table}[h!]
\centering
\caption{Mean of mixed-Hamming error rates for ego-networks.}
\label{ErrorSNAP}
\begin{tabular}{cccccccccc}
\toprule
&Facebook&GooglePlus&Twitter\\
\midrule
Mixed-SCORE&\textbf{0.2496}&0.3766&0.3087\\
OCCAM&0.2610&0.3564&0.2863\\
GeoNMF&0.2584&0.3507&0.2859\\
\hline
Mixed-SLIM&0.2521&0.3105&0.2706\\
Mixed-$\mathrm{SLIM}_{\tau}$&0.2507&0.3091&0.2656\\
Mixed-$\mathrm{SLIM}_{appro}$&0.2516&0.3161&0.2694\\
Mixed-$\mathrm{SLIM}_{\tau appro}$&0.2504&\textbf{0.3088}&\textbf{0.2624}\\
\bottomrule
\end{tabular}
\end{table}
\begin{figure*}
	\vskip 0.2in
\centering
\subfigure[Facebook]{\includegraphics[width=0.3\textwidth]{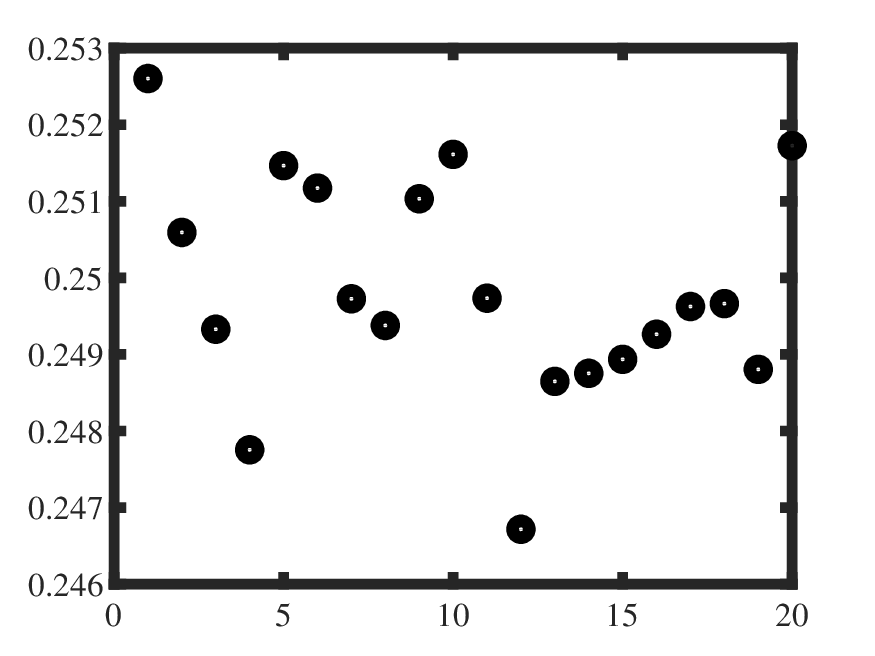}}
\subfigure[GooglePlus]{\includegraphics[width=0.3\textwidth]{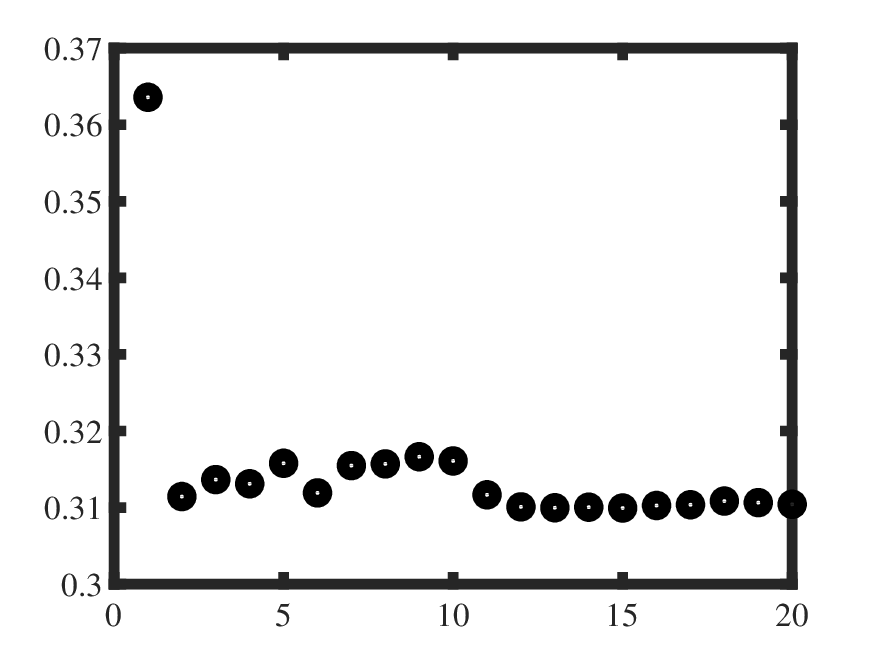}}
\subfigure[Twitter]{\includegraphics[width=0.3\textwidth]{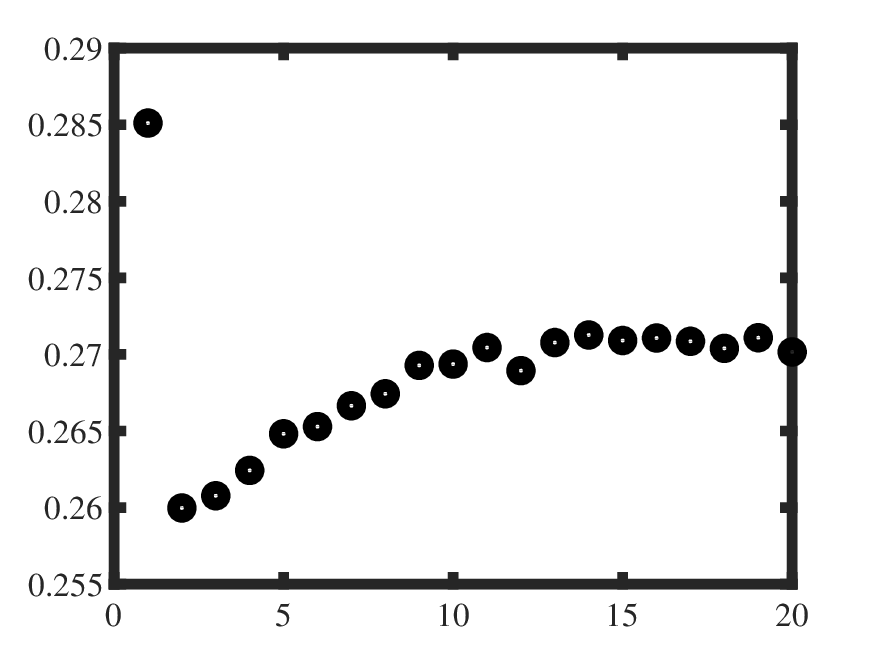}}
\caption{Mean of mixed-Hamming error rates on SNAP ego-networks for Mixed-$\mathrm{SLIM}_{appro}$ based on different choice of $T$, where $T$ is in $\{1,2, \ldots, 20\}$. x-axis: $T$, y-axis: mean of mixed-Hamming error rates.}
\label{MixedSLIMapproTSNAP} 
\vskip -0.2in
\end{figure*}
\subsection{Discussion on the choice of $T$}\label{StudyT}
We study the effect of $T$ to Mixed-$\mathrm{SLIM}_{appro}$ by changing $T$ from 1 to 20 on SNAP datasets. The numerical results are shown in Figures  \ref{MixedSLIMapproTSNAP}. These results suggest that Mixed--$\mathrm{SLIM}_{appro}$ is insensitive to the choice of $T$ as long as it is larger than or equal to 3. Meanwhile, as discussed in \cite{SLIM}, when $T$ is too large, $\hat{D}^{-T}A^{T}$ no longer contains any block structure, and all nodes merge into one giant community since $\hat{D}^{-T}$ is nearly singular for large $T$. Therefore the ideal choice for $T$ should not be too large in practice.  We set the default value for $T$ as 10 in this article. 
\section{Discussion}\label{sec6}
This paper makes one major contribution: modified SLIM methods to mixed membership community detection under the DCMM model.   When dealing with large networks in practice, we apply Mixed-$\mathrm{SLIM}_{appro}$ and its regularized version Mixed-$\mathrm{SLIM}_{\tau appro}$. We showed the estimation consistency of the regularized version Mixed-$\mathrm{SLIM}_{\tau}$ under the DCMM model. Both simulation and empirical results for community detection and mixed membership community detection demonstrate that Mixed-SLIM methods enjoy satisfactory performances and they perform better than most of the benchmark methods.

Our idea proposed in this paper can be extended in many directions. Building the theoretical framework for Mixed-$\mathrm{SLIM}_{appro}$ and Mixed-$\mathrm{SLIM}_{\tau appro}$ is challenging and interesting, and we leave it as future work. In Mixed-SLIM methods, the SLIM matrix is computed by $(I-\alpha \hat{D}_{\tau}^{-1}A_{\tau})^{-1}$, we wonder that whether there exists an optimal parameter $\beta_{0}$ such that Mixed-SLIM methods designed based on the new SLIM matrix $(I-\alpha \hat{D}_{\tau}^{-\beta_{0}}A_{\tau})^{-1}$ outperform those designed based on $(I-\alpha \hat{D}_{\tau}^{-\beta}A_{\tau})^{-1}$ for any $\beta$ both theoretically and empirically. Similar to \cite{su2019strong}, it is also interesting to build a theoretical guarantee of strong consistency of spectral algorithms designed based on SLIM under DCSBM and DCMM. Meanwhile, it remains unclear how to estimate the number of communities K for the network with mixed memberships, and we wonder whether the SLIM matrix and its regularized one can be applied for estimating $K$ both theoretically and empirically under MMSB and DCMM. Furthermore, \cite{rohe2016co} extends SBM and DCSBM from un-directed networks to directed networks. We wonder whether the idea of SLIM for un-directed networks can be extended to directed networks. For reasons of space, we leave studies of these problems to the future.

\backmatter

\bmhead{Supplementary information}

The proofs of lemmas and theorems are provided in the Supplementary.

%

\section*{Declarations}


\begin{itemize}
	\item Funding. Qing's work was supported by High level personal project of Jiangsu Province (JSSCBS20211218). Wang’s work was supported by the Fundamental Research Funds for the Central Universities, Nankai University, 63221044 and the National Natural Science Foundation of China (Grant 12001295).
	\item Conflict of interest/Competing interests (check journal-specific guidelines for which heading to use) None
	\item Ethics approval None
	\item Consent to participate YES
	\item Consent for publication YES
	\item Availability of data and materials YES
	\item Code availability YES
	\item Authors' contributions. Qing mainly worked on the algorithm and theoretical properties. Wang mainly worked on the algorithm and whole paper organization.
\end{itemize}
\bibliographystyle{sn-basic}
\bibliography{refMixedSLIM}

\end{document}